\documentclass{article}

     \usepackage[preprint]{neurips_2024}

\usepackage[utf8]{inputenc} 
\usepackage[T1]{fontenc}    
\usepackage{hyperref}       
\usepackage{url}            
\usepackage{booktabs}       
\usepackage{amsfonts}       
\usepackage{nicefrac}       
\usepackage{microtype}      
\usepackage{xcolor}         
\usepackage{amsmath}
\usepackage{amssymb}
\usepackage{mathtools}
\usepackage{amsthm}
\theoremstyle{plain}

\theoremstyle{definition}

\theoremstyle{remark}

\usepackage{hyperref}
\usepackage{algorithm}
\usepackage{algorithmic}

\title{A Bayesian explanation of machine learning models based on modes and functional ANOVA }

\author{%
  Quan.~Long\thanks{Use footnote for providing further information  about author (webpage, alternative address)---\emph{not} for acknowledging
    funding agencies.} \\
  Honeywell International\\
  Northford, Connecticut, USA \\
  \texttt{quan.spartanlq@gmail.com} \\
}

\begin{document}

\maketitle

\begin{abstract}
Most methods in explainable AI (XAI) focus on providing reasons for the prediction of a given set of features. However, we solve an inverse explanation problem, i.e., given the deviation of a label, find the reasons of this deviation. We use a Bayesian framework to recover the ``true'' features, conditioned on the observed label value. We efficiently explain the deviation of a label value from the mode, by identifying and ranking the influential features using the ``distances'' in the ANOVA functional decomposition. We show that the new method is more human-intuitive and robust than methods based on mean values, e.g., SHapley Additive exPlanations (SHAP values). The extra costs of solving a Bayesian inverse problem are dimension-independent.
\end{abstract}

\section{Introduction}
Explaining the deviation of an observation is of great practical importance. For example, many industrial suppliers' on-time delivery rates significantly dropped in 2021 due to the COVID-19 associated factors, such as lock-downs and lack of labors. The productivity of a manufacturing facility may deviate from its calibrated value due to aging of equipment, unexpected shifts of the controls and abnormal climates. Which factors and to what extends they contribute to the change of the output are among the typical questions a decision maker would inquire about. 

Recent explainable AI (XAI) has been focusing on explaining a forward model, i.e., methods have been developed to estimate the impacts of the features on a prediction. The local interpretable model-agnostic explanation (LIME) developed by \cite{LIME} constructs a linear local model whose coefficients explicitly explain the importance of the corresponding features. Hence, it is only valid in the region where the linear model is a good approximation of the original model. The SHAP values (see the works of  \cite{RePEc:wly:apsmbi:v:17:y:2001:i:4:p:319-330, NIPS2017_8a20a862, JMLR:v24:22-0202} for details) provide scores indicating the importance of each feature. The scores measure the differences between the predictions of the model trained by the whole dataset and the predictions of the model trained on a sub-dataset without the subset of the respective features. \cite{Giles:2007} uses ANOVA functional decomposition of a model for explanation. It represents another important stream of methods, which stem from statistical design of experiments. \cite{Giles:2007} also shows both the responsible scores in SHAP and the terms in ANOVA are combinations of conditional expectations of the original model. Analyses in the recent literature, e.g., studies by \cite{owen:SISV} and \cite{herren2022statistical}, show that ANOVA provides sharp bounds for the SHAP values and are identical to them up to a proportional constant for linear models.  

Bayes theorem provides a systematic framework to compute a conditional distribution by using prior distribution and a likelihood function. Much has been written about the applications of Bayes framework in AI and machine learning (e.g., \cite{NIPS2017_64a08e5f};  \cite{gal2017deep}). Recently, a likelihood based inverse explainable AI method has been proposed in \cite{Ide_Dhurandhar_Navratil_Singh_Abe_2021}. In this work, the authors use likelihood function to inversely explain a deviation of a label in the scenario, where the training data are not available and the model is a complex black box. 

In our work, we follow the Bayesian framework to find the ``true'' features and use the ANOVA functional decomposition to explain the contributions of the features towards the deviation of the label from a reference value. Importantly, instead of using sample-mean-based approach, we use the modes. The mode-based approach provides salient robustness, while the mean value may be influenced by outliers on explaining complex black-box relationships, e.g., multimodality. We use a direct search method developed in \citep{Quan_Long_2022} to find the Bayesian maximum a posteriori estimate (MAP) of the label mode. The direct search solves a series of optimization problems with a dimension-independent computational cost.

%
%
%
%
%

The layout of this paper is described as follows: section \ref{sec:bayes} describes the Bayesian framework of inversely explaining a label value; section \ref{sec:decom} introduces the functional decomposition of a deviation from a mode; section \ref{sec:scores} defines the mode specific responsible scores; in section \ref{sec:examples}, two examples are used to illustrate the accuracy of the proposed method. In the first example, we show that the proposed scoring scheme produces feature importance which is more intuitive than the mean-value-based SHAP values. In the second example, the proposed approach is used to explain the measurements of water flow in a major river in England. The proposed approach is shown to be more robust than an outlier explainer using SHAP values.

\section{Bayesian explanation of a label value}\label{sec:bayes}
We use the following form of Bayes theorem to inversely compute the distributions of the features, conditioned on the observed labels: 
\begin{align}
p(\boldsymbol{x}_i | y_i) = \frac{p(y_i|\boldsymbol{x}_i) p(\boldsymbol{x}_i)}{p(y_i)} \,, \quad i = 1,...,N \nonumber
\end{align}
where $p(\cdot)$ is a probability density function (pdf), $p(\boldsymbol{x}_i|y_i)$ is the posterior pdf of feature $\boldsymbol{x}_i \in \boldsymbol{R}^{d_x}$ given the label $y_i \in \boldsymbol{R}$, $p(y_i|\boldsymbol{x}_i)$ is the likelihood function, $p(\boldsymbol{x}_i)$ is the prior pdf of feature $\boldsymbol{x}_i$, $p(y_i) = \int_{\boldsymbol{x}_i\in {\Omega}} p(y_i|\boldsymbol{x}_i)p(\boldsymbol{x}_i)d\boldsymbol{x}_i $ is a normalization constant, $\Omega$ is the parametric space of $\boldsymbol{x}$, $N$ is the number of samples - pairs of $\boldsymbol{x}_i$, and $y_i$ in a dataset. The likelihood function can be further written as follows: 
\begin{align}
p(y_i|\boldsymbol{x}_i) = \frac{1}{(2\pi \prod_{i=1,...,N}\sigma^2_i)^{1/2}}e^{-\sum_{i=1}^N[y_i - f(\mathbf{x}_i,\boldsymbol{\theta}^*)]^2/\sigma^2_i} \,, \nonumber
\end{align}  
where $f(\mathbf{x},{\boldsymbol{\theta}}^*)$ is a trained machine learning model, e.g., linear regression, extreme gradient boosting, neural network, etc. In a typical training process, optimization is used to obtain $\boldsymbol{\theta}^*$, e.g., the mean square error is minimized to obtain the coefficients of a regression model as below:  
\[\boldsymbol{\theta}^* = \arg\min_{\boldsymbol{\theta} \in {\cal{\boldsymbol{\Theta}}}} \sum^N_{i=1}[y_i - f(\mathbf{x}_i, \boldsymbol{\theta})]^2 \,, \]
where $\boldsymbol{\theta}$ is the vector of the coefficients in the regression model, ${\cal{\boldsymbol{\Theta}}}$ is the parametric space. 
Using the fore-mentioned Bayesian approach, we can solve for the ``true'' values of the features, namely the maximum a posteriori (MAP) estimate of the features, given the label $y_i$:
\begin{align}
\hat{\boldsymbol{x}}_i = \arg\max_{\boldsymbol{x}_ \in \Omega} log[p(\boldsymbol{x}_i|y_i)]\,. \label{eq:map_mode}
\end{align}
For the sake of conciseness, we will neglect $\boldsymbol{\theta}^*$ in $f(\mathbf{x}_i, \boldsymbol{\theta})$ in the rest of the paper. 
\section{Explaining a label's deviation}\label{sec:decom}
Explaining the deviation of a label from a reference is a rarely investigated topic, except several recent attempts on outlier explanations, e.g. by \cite{Ide_Dhurandhar_Navratil_Singh_Abe_2021}, where the authors treat the probability estimates of the rare events as the quantity of interest and compute the associated SHAP value.  We firstly focus on explaining the deviation of the $i^{th}$ label from its mean value. This deviation can be decomposed into a series of ANOVA functions as the following:
\begin{align}
\delta_i = y_i - \bar{y} =f(\boldsymbol{x}_i) - f(\boldsymbol{x}') +\epsilon_i=& \sum_{I=1}^{d_x} \delta_{iI} 
+ \sum_{I<J}^{d_x} \delta_{iIJ} + \sum_{I<J<K}^{d_x} \delta_{iIJK}+ ... + \epsilon_i \,, \quad i = 1,...,N \label{eq:delta}
\end{align}
with
\begin{align}
\delta_{iI} = & f_I (x_{iI}) - f_I(x'_{I})\,, \nonumber\\
\delta_{iIJ} = & f_{IJ} (x_{iI}, x_{iJ}) - f_{IJ}(x'_{I},x'_{J}) \,,\nonumber\\
\delta_{iIJK} = & f_{IJK}(x_{iI}, x_{iJ}, x_{iK}) - f_{IJK}(x'_{I}, x'_{J}, x'_{K})\,,\nonumber
\end{align}
where $x_{iI}$ denotes the $I^{th}$ feature of the $i^{th}$ data point.
The functions $f_I(\cdot)$, $f_{IJ}(\cdot,\cdot)$, $f_{IJK}(\cdot,\cdot,\cdot)$ are the ANOVA functional decompositions of the machine learning model $f$.
Appendix A shows the definition of these functions. Appendix B shows how they are estimated using Monte Carlo sampling. Recall that we can obtain the ``true'' feature values by solving a maximization problem of  \eqref{eq:map_mode}. Similarly, we can solve a maximization problem to obtain $\boldsymbol{x}'$ as below: 
 
\begin{align}
\mathbf{x}' = \arg\max_{\mathbf{x}\in \Omega}  log \left[p( \bar{y} | \mathbf{x})\right] + log\left[p(\mathbf{x})\right] \,. \label{eq:mapmean}
\end{align}
Note that $\boldsymbol{x}'$ is not necessarily the sample mean of the features, unless the model $f$ is linear.

\subsection{Mode-specific deviation}
In this section, we introduce mode specific deviations, because the deviation from mean value is non-intuitive in many scenarios, e.g., the dataset has multiple modes, or the mode and mean values differ significantly. Replacing $\bar{y}$ in Equation \eqref{eq:delta} by a mode leads to the following definition of mode-specific deviation:
\begin{align}
\delta^m_i = y_i - y^*_m = f(\mathbf{x}_i) - f(\mathbf{x}^*_m)+ \epsilon_i \,, \nonumber
\end{align}
where $y^*_m$ is the $m^{th}$ mode. Consequently, we decompose the above deviation using ANOVA functional decomposition as follows:
\begin{align}
f(\boldsymbol{x}_i ) - f(\boldsymbol{x}^*_m) + \epsilon_i =& \sum^{d_x}_{I=1} \delta_{iI}^m 
+ \sum^{d_x}_{I<J} \delta_{iIJ}^m 
+ \sum^{d_x}_{I<J<K} \delta_{iIJK}^m + ... 
+ \epsilon_i \,, \nonumber
\end{align}
with
\begin{align}
\delta_{iI}^m = & f_I (x_{iI}) - f_I(x_{mI}^*)\,, \nonumber\\
\delta_{iIJ}^m = & f_{I,J} (x_{iI}, x_{iJ}) - f_{I,J}(x_{mI}^*, x_{mJ}^*) \,, \nonumber\\
\delta_{iIJK}^m = & f_{I,J,K}(x_{iI}, x_{iJ}, x_{iK}) - f_{I,J,K}(x_{mI}^*, x_{mJ}^*, x_{mK}^*) \,, \nonumber
\end{align}
and
\begin{align}
\mathbf{x}^*_m = \arg\max_{\mathbf{x}\in \Omega}  log \left(p( y^*_m | \mathbf{x})\right) + log\left(p(\mathbf{x})\right) \,,\label{eq:modInv}
\end{align}
where the log likelihood function is
\begin{align}
log \left(p( y^*_m | \mathbf{x})\right) \propto - \frac{\left[y^*_m - f(\mathbf{x})\right]^2}{2\sigma_e^2} \,,\nonumber
\end{align}
and 
\begin{align}
\sigma^2_e = \frac{\sum^{N}_{i=1}(f(\mathbf{x}_i) - y_i)^2}{N}\,. \nonumber
\end{align}

\subsection{Direct search of the MAP}
\eqref{eq:map_mode}, \eqref{eq:modInv} and \eqref{eq:mapmean} and  may have multiple local optimums. Nevertheless, we use several runs of optimization to find a collection of local optimums of the maximization problems. Then, we can obtain the MAP in the set of the local optimums. The steps are described in Algorithm \ref{alg:map} as shown above. 

\begin{algorithm}
\caption{Direct search of MAP in \cite{Quan_Long_2022}}\label{alg:map}
\begin{algorithmic}[1]
\REQUIRE $p(\mathbf{x}|\mathbf{y})$, $f(\mathbf{x})$, $no$
\STATE draw a sample from the distribution 
\STATE solve the maximization problem \eqref{eq:map_mode} $no$ times with distinct initial points, the distinct local optimal solutions are 
$\{\mathbf{x}^*_k, k = 1,...,K\}$
\STATE $\boldsymbol{x}^*_m = \arg\max_{k = 1,...,K} log \left(p( y^*_m | \mathbf{x}_i)\right) + log\left(p(\mathbf{x}_i)\right) $
\end{algorithmic}
\end{algorithm}

\cite{Quan_Long_2022} shown that the number of necessary runs is a dimension-independent value bounded by the inequality below: 
\begin{align}
no \geq \frac{\log \beta - \log K}{log (1-p)}\,,\nonumber
\end{align}
where $K$ is the number of local optimums, $p$ is the minimum probability that an independent and identically distributed (i.i.d.) initial value of $\boldsymbol{x}$ converges to a local optimum, $\beta$ is the probability that $no$ runs of optimization do not find all the local optimums. 

\section{Mode-specific responsible scores}\label{sec:scores}
We define the responsible scores, which are the contributions to $\delta^m_i$ by a feature or a subset of the features:
\begin{align}
s_{iI}^m = \frac{\delta_{iI}^m}{\delta^m_i}\,,\quad\nonumber
s_{iIJ}^m = \frac{\delta_{iIJ}^m}{\delta^m_i}\,,\quad \nonumber
s_{iIJK}^m = \frac{\delta_{iIJK}^m}{\delta^m_i}\,. \nonumber
\end{align}

Note that high order interactions are usually negligible compared to the first order responsible scores \citep{owen:SISV}. In the next section, we use the mode-specific responsible scores to identify and rank the input features, which contribute to the deviation of the output label against its mode.

\section{Experiments}\label{sec:examples}
\subsection{Outlier reasoning using synthetic data}

\begin{figure}[ht]
\begin{center}
\centerline{\includegraphics[scale=0.4]{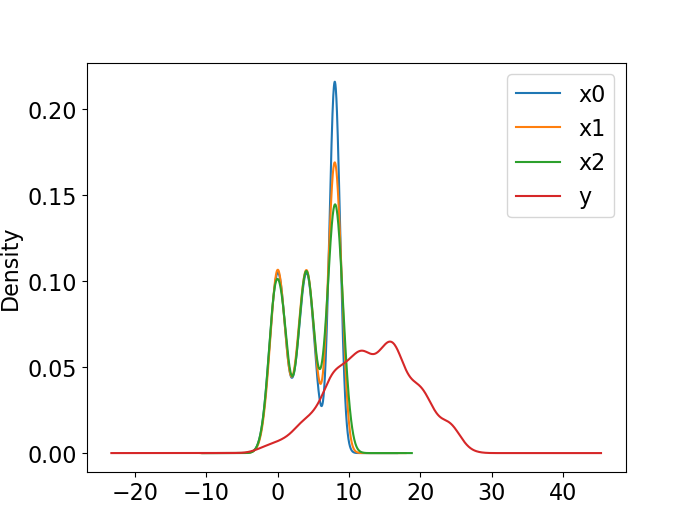}}
\caption{Distributions of the multimodal dataset, where $y=x_0 + x_1 + x_2$.}
\label{fig:dist_multimodal}
\end{center}
\vskip -0.2in
\end{figure}

In this example we first draw i.i.d. samples of $x_0$, $x_1$ and $x_2$. 
$x_0$ is drawn from the Gaussian mixture distribution $0.3{\cal{N}}(0, 1) + 0.3{\cal{N}}(4, 1) + 0.4{\cal{N}}(8, 0.5^2)$. 
$x_1$ is drawn from the Gaussian mixture distribution $0.3{\cal{N}}(0, 1) + 0.3{\cal{N}}(4, 1) + 0.4{\cal{N}}(8, 0.75^2)$.
$x_2$ is drawn from the Gaussian mixture distribution $0.3{\cal{N}}(0, 1) + 0.3{\cal{N}}(4, 1) + 0.4{\cal{N}}(8, 1)$.
They are independent to each other. We use $y = x_0 + x_1 + x_2$ to synthetically generate the samples of the dependent variable $y$. 
The marginal distributions of $x_0$, $x_1$, $x_2$ and $y$ are visualized in Figure \ref{fig:dist_multimodal}. 

In total, $10000$ synthetic data points are generated. A Gaussian mixture based clustering method by \cite{NIPS2017_dfeb9598} is used to find the clusters in the set of the labels. The dominant mode is $y^* = 15.7$.
The mean value is $\bar{y} = 13.2$. 

The smallest $y$ in the sample is $-6.2$. Its z-score and mode-based z-score $(z_m = \frac{\delta_m}{\sigma_m})$ are listed in Table \ref{tab:zscorey}. We define the $m^{th}$ mode based z-score as the distance from the sample to the $m^{th}$ mode, normalized by $\sigma_m$- the standard deviation of the $m^{th}$ mode of the approximated Gaussian mixture. The independent variables corresponding to the outlier of $y=-6.2$ are $x_0 = -2.5$, $x_1 = -1.7 $ and $x_2 = -2.0$. 
\begin{table}[ht]
\begin{center}
\begin{tabular}{c c c }
\hline
Outlier & $z$ & $z_m$  \\
\hline
$-6.2$ & $-3.3$ & $-15.6$ \\
\hline
\end{tabular}
\end{center}
\caption{The smallest $y$ and its $z$-scores w.r.t. the sample mean and the mode. }\label{tab:zscorey}
\end{table}

A linear regression model is used to fit the $10000$ points in the sample. We use $p(\boldsymbol{x})=p(x_1)p(x_2)p(x_3)$, where $p(x_1)$, $p(x_2)$ and $p(x_3)$ are the fore-mentioned Gaussian mixtures.
We then solve the maximization problem \eqref{eq:modInv} for $y_m^*=15.7$, using the Broyden–Fletcher–Goldfarb–Shanno (BFGS) method described in \cite{pmlr-v51-moritz16}. The MAP estimate of $x^*_m$ is $[7.97, 7.94, -0.11]$.

\begin{figure} [ht]
\begin{center}
\centerline{\includegraphics[scale=0.35]{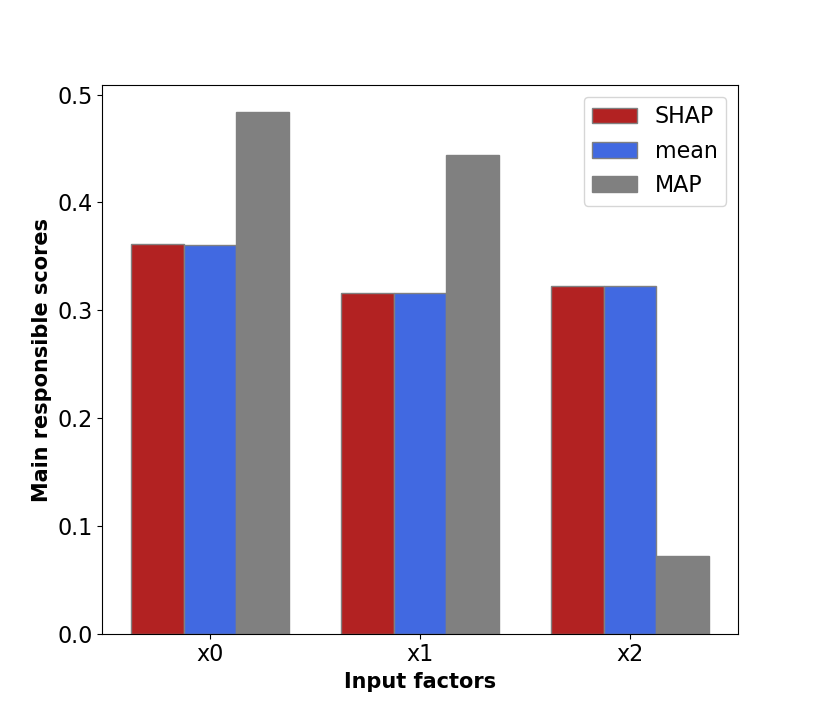}}
\caption{The SHAP values and the main responsible scores of $x_0$, $x_1$ and $x_2$ explaining $y$'s deviation from the mean and the dominant mode.}
\label{fig:multimodal_bar_charts}
\end{center}
\vskip -0.2in
\end{figure}

 Figure \ref{fig:multimodal_bar_charts} shows the main responsible scores based on the mean value and the dominant mode of the label $y$, as well as the SHAP values. The scores based on mean value equal the SHAP values due to the linearity of the model. This is theoretically demonstrated by \cite{owen:SISV} and \cite{herren2022statistical}. Note that $x_0$ contributes the most and $x_1$ contributes the least to $y$'s largest deviation from the mean value. Overall, the three input factors contribute slightly above $30\%$ respectively. To the contrary, $x_0$ contributes the most and $x_2$ contributes the least to $y$'s deviation against the mode. Specifically, $x_0$ contributes to $48\%$, $x_1$ contributes to $45\%$ and $x_2$ contributes to $7\%$. 
Because of the multimodal distribution of $\boldsymbol{x}$ and $y$, a deviation to the mode is more human-intuitive than a deviation to the mean value. We demonstrate that the rankings of the input variables are significantly different in both scenarios. 




\subsection{Explaining seasonal flows of a river}
In this experiment, we use our approach to explain the river flow measured at the New Jumbles Rock station ($njr$) in England. 
There are three upstream rivers, namely Hodder Place ($hp$), Henthorn ($h$) and Whalley Weir ($ww$). This dataset 
was also used in \cite{Ide_Dhurandhar_Navratil_Singh_Abe_2021} for root-cause analysis. 
River flows have been measured in each station at 9:00 am daily, since 1
January 2010. We use the daily river flow at njr as the label, and the daily river flows at  $p$ , $h$ and $ww$ as the features to train a machine learning model. 

We first use a linear regression model to fit the dataset. The $R^2=98.9\%$ in predicting the training labels, while $R^2=99.2\%$ in predicting the testing labels. The estimated 
coefficients are $1.2$, $0.6$ and $1.02$ with respect to $ru_h$, $ru_{hp}$ and $ru_{ww}$ ($ru$ indicate upstream rivers). 

\begin{figure}[ht]
\begin{center}
\centerline{\includegraphics[scale=0.5]{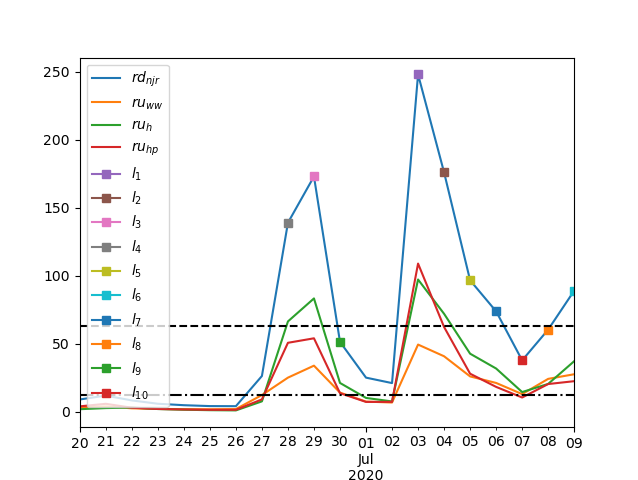}}
\caption{Time histories of river flow measurements.}
\label{fig:timehist}
\end{center}
\vskip -0.2in
\end{figure}

We then use our approach described in Sections 3 and 4 to explain the flow's deviations in the time range between June 20, 2020 and July 09, 2020. Figure 
\ref{fig:timehist} shows the time histories of the river flow measurements. The mode of the downstream flow is $12.2$ marked by the dash-dotted line in Figure \ref{fig:timehist}. The corresponding MAP estimate of the upstream flows are $3.7$, $7.4$, and $2.6$ for $ru_h$, $ru_{hp}$ and $ru_{ww}$, respectively. The mean values of the downstream flow is $63.4$, denoted by the dashed line in Figure \ref{fig:timehist}. The mean values of the upstream flows are $26.4$, $21.7$ and $16.1$ for $ru_h$, $ru_{hp}$ and $ru_{ww}$, respectively. We highlight the ten largest $rd_{njr}$ ($rd$ means downstream river) river flow measurements, denoted by $l_1, ..., l_{10}$ in Figure \ref{fig:timehist}.

%

\begin{figure}[ht!]
\begin{center}
\includegraphics[scale=0.4]{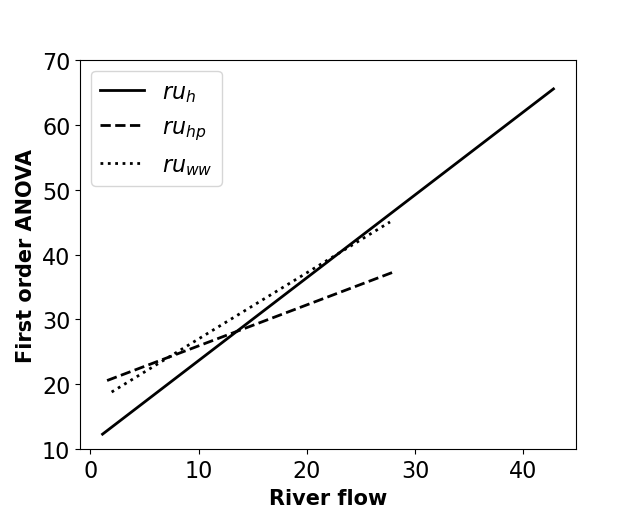}
\includegraphics[scale=0.4]{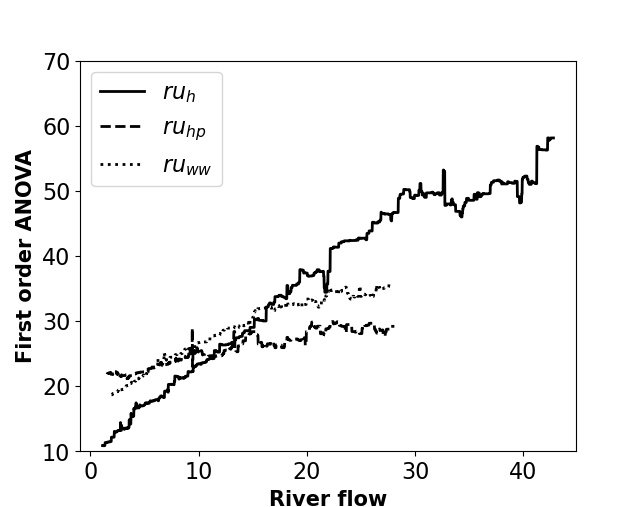}
\caption{The first order ANOVA functions of the river flows of $ru_h$, $ru_{hp}$ and $ru_{ww}$, modeled by linear regression on the left and extreme gradient boosting on the right.}
\label{fig:firstANOVA_lr}
\end{center}
\vskip -0.2in
\end{figure}

\begin{figure}[ht!]
\begin{center}
\includegraphics[scale=0.5]{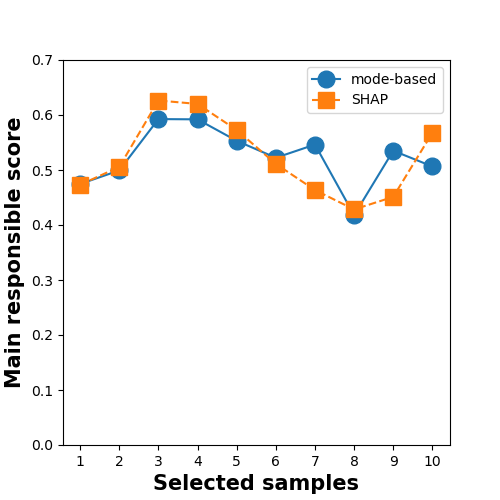}
\includegraphics[scale=0.5]{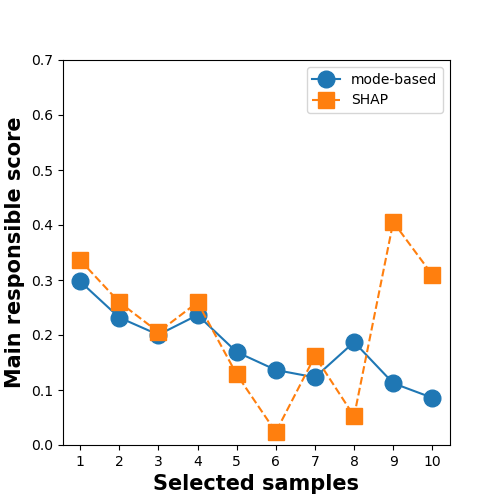}
\includegraphics[scale=0.5]{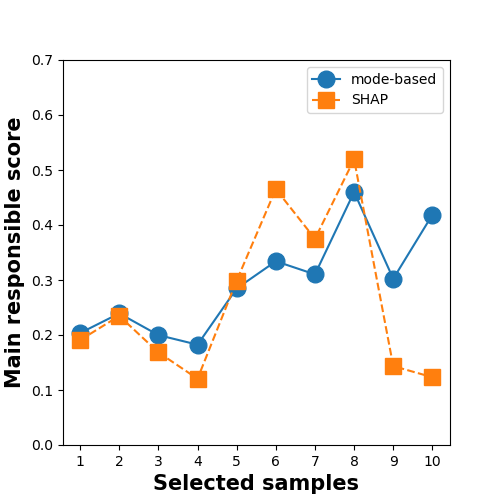}
\caption{Main responsible scores of $ru_h$ (upper-left), $ru_{hp}$ (upper-right) and $ru_{ww}$ (bottom).}
\label{fig:map_scores_rivers}
\end{center}
\vskip -0.2in
\end{figure}

The first order ANOVA functions are plotted on the left of Figure \ref{fig:firstANOVA_lr}. The slopes of the linear ANOVA functions match the coefficients of the model. 
The main responsible scores computed based on the MAP of the modes and the SHAP values for ten sample points ($l_1$, $l_2$, ..., $l_{10}$) are plotted in Figures \ref{fig:map_scores_rivers}. It is observed that the mode-based approach and the SHAP values are close to each other for samples 1-5. However, they differ significantly for samples 6-10. 

\begin{figure}[ht!]
\begin{center}
\includegraphics[scale=0.55]{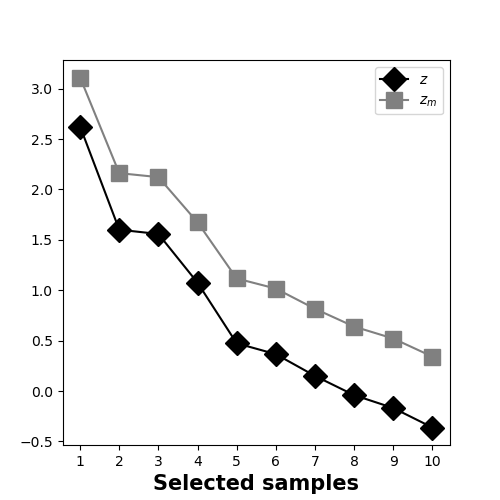}
\caption{The $z$ and $z_m$ of ten selected samples of $ru_{njr}$. They are used to measure a sample's distance to the mean and the mode, respectively.}
\label{fig:z_scores}
\end{center}
\vskip -0.2in
\end{figure}

It is observed that the flow from $ru_h$ is always the dominant contributor to the downstream river, while the ranks of $ru_{hp}$ and $ru_{ww}$ depend 
on the dates. When the river flows of $ru_{hp}$ and $ru_{ww}$ are close to the temporal mean values (see the $9$-th and $10$-th samples in upper-right and bottom snapshots of Figure \ref{fig:map_scores_rivers}), the SHAP values get unstable, while the trends of the mode-based responsible scores are stable. Figure \ref{fig:z_scores} shows the mean and mode-based z-scores of the ten samples. Note that the $z_m$ are larger than the $z$ by $0.5$ consistently.



We then use extreme gradient boosting ( \cite{Chen:2016}) to fit the dataset. The R-squared value $R^2=99.6\%$ in predicting the training labels, and the R-squared value $R^2=99.7\%$ in predicting the testing labels. The first order ANOVA functions are plotted on the right of Figure \ref{fig:firstANOVA_lr}. The noise comes from the Monte Carlo sampling in Equations \eqref{eq:deltai} and the non-smoothness of the extreme gradient boosting model.

\begin{figure}[ht!]
\begin{center}
\includegraphics[scale=0.45]{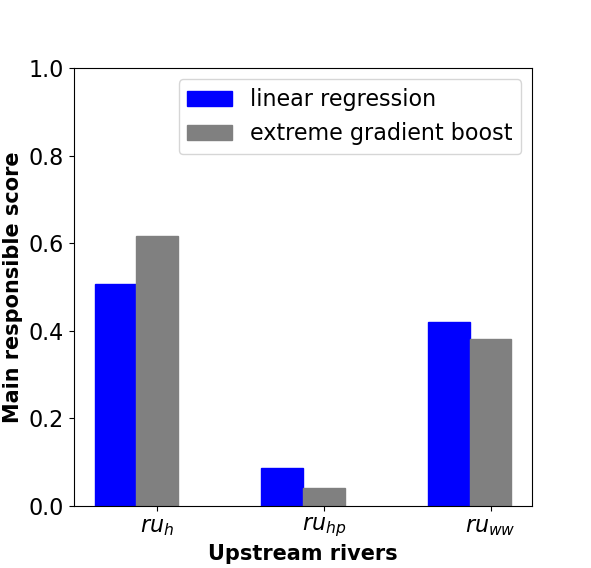}
\includegraphics[scale=0.45]{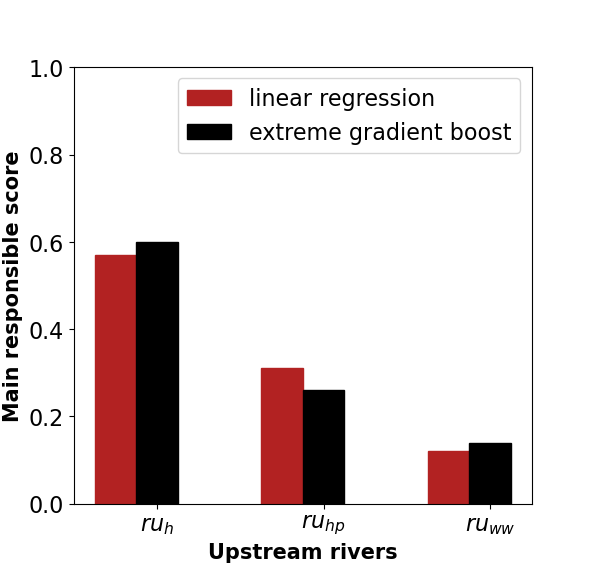}
\caption{The mode-based responsible scores (left) and SHAP values (right) of $ru_h$, $ru_{hp}$ and $ru_{ww}$.}
\label{fig:map_point10}
\end{center}
\vskip -0.2in
\end{figure}

The bar charts in Figure \ref{fig:map_point10} show the main responsible scores of the three upstream rivers for $l_{10}$. The relationships between upstream and downstream river flows are modeled using linear regression and extreme gradient boosting. The results obtained using both models are mostly consistent. Hence, our method is shown to be model-agnostic.

%
%

\section{Conclusion}
We use a systematic Bayesian framework to explain deviations of an AI model from its mode. The contribution of the features, namely the responsible scores, are defined as the distances between the Bayesian MAPs in the ANOVA functional subspace. We extend the conventional mean-value-based scores to mode-based scores, which is more robust and human-intuitive. The new approach is tested using two experiments. The first example shows that the mode-based method provides better human-intuitive results than the mean-value-based methods in a multimodal scenario. In the second experiment, we use the proposed approach to explain the flow outliers of a major river in England. The approach 
is able to explain the importance of the features in terms of their corresponding  
functional decompositions of the AI models. The mode-based main responsible scores show salient stability in the vicinity of mean values, while existing method becomes unstable.

\section{Appendix}
\subsection{Function decomposition}

The $f(\boldsymbol{x})$ can be decomposed using ANOVA representation as follows (see  \cite{SOBOL2001271,SALTELLI2010259} for details)

\begin{align}
f(\boldsymbol{x}_i) =& f_0 + \sum^{d_x}_{I=1} f_{I} (x_{iI}) + \sum^{d_x}_{I<J} f_{IJ} (x_{iI}, x_{iJ}) + \sum^{d_x}_{I<J<K} f_{IJK}(x_{iI}, x_{iJ}, x_{iK})+ ... + f_{12...d_x}(x_{i1}, x_{i2}, ..., x_{id_x})\,,\nonumber
\end{align}
where the terms of the RHS are the expectations of $f$ conditioned on an increasing subset of the random inputs $\boldsymbol{x}$. Specifically, the first order and second order functions can be written as follows: 
\begin{align}
f_I(x_{iI}) =& \mathbf{E}_{\mathbf{x}_i \backslash x_{iI}}(f | x_I) - f_0 \,,\nonumber\\
f_{IJ}(x_{iI}, x_{iJ}) =& \mathbf{E}_{\mathbf{x}_i \backslash (x_{iI}, x_{iJ})}(f | x_I, x_J) - f_I(x_{iI}) - f_J(x_{iJ}) - f_0\,.\nonumber
\end{align}

\subsection{Estimation of the conditional functions using Monte Carlo}

The first order ANOVA of $\delta_i$ can be estimated numerically using random samples.

\begin{align}
\delta_{iI} = & \mathbf{E}_{\mathbf{x} \backslash \boldsymbol{x}_{iI}}(f | \boldsymbol{x}_{iI}) - \mathbf{E}_{\mathbf{x} \backslash \boldsymbol{x}_{iI}}(\bar{f} | \boldsymbol{x}_{iI}) 
    =  \frac{1}{NP} \sum^{NP}_{s=1} f( \mathbf{x}_s \backslash \boldsymbol{x}_{iI} )- \frac{1}{NP} \sum^{NP}_{s=1} f( {\boldsymbol{x}}'_s \backslash {\boldsymbol{x}}'_{iI} ) + \mathcal{O}_P(\frac{1}{\sqrt{NP}})\,, \label{eq:deltai}
\end{align}


\begin{thebibliography}{10}
\providecommand{\natexlab}[1]{#1}
\providecommand{\url}[1]{\texttt{#1}}
\expandafter\ifx\csname urlstyle\endcsname\relax
  \providecommand{\doi}[1]{doi: #1}\else
  \providecommand{\doi}{doi: \begingroup \urlstyle{rm}\Url}\fi

\bibitem[Chaloner \& Verdinelli(1995)Chaloner and
  Verdinelli]{chaloner1995bayesian}
Chaloner, K. and Verdinelli, I.
\newblock {B}ayesian experimental design: A review.
\newblock \emph{Statistical science}, pp.\  273--304, 1995.

\bibitem[Chen \& Guestrin(2016)Chen and Guestrin]{Chen:2016}
Chen, T. and Guestrin, C.
\newblock {XGBoost}: A scalable tree boosting system.
\newblock In \emph{Proceedings of the 22nd ACM SIGKDD International Conference
  on Knowledge Discovery and Data Mining}, KDD '16, pp.\  785--794, New York,
  NY, USA, 2016. ACM.
\newblock ISBN 978-1-4503-4232-2.

\bibitem[Gal et~al.(2017)Gal, Islam, and Ghahramani]{gal2017deep}
Gal, Y., Islam, R., and Ghahramani, Z.
\newblock Deep {B}ayesian active learning with image data.
\newblock In \emph{International conference on machine learning}, pp.\
  1183--1192. PMLR, 2017.

\bibitem[Herren \& Hahn(2022)Herren and Hahn]{herren2022statistical}
Herren, A. and Hahn, P.~R.
\newblock Statistical aspects of shap: Functional anova for model
  interpretation, 2022.

\bibitem[Hooker(2007)]{Giles:2007}
Hooker, G.
\newblock Generalized functional anova diagnostics for high-dimensional
  functions of dependent variables.
\newblock \emph{Journal of Computational and Graphical Statistics}, 16\penalty0
  (3):\penalty0 709--732, 2007.

\bibitem[Idé et~al.(2021)Idé, Dhurandhar, Navrátil, Singh, and
  Abe]{Ide_Dhurandhar_Navratil_Singh_Abe_2021}
Idé, T., Dhurandhar, A., Navrátil, J., Singh, M., and Abe, N.
\newblock Anomaly attribution with likelihood compensation.
\newblock \emph{Proceedings of the AAAI Conference on Artificial Intelligence},
  35\penalty0 (5):\penalty0 4131--4138, May 2021.

\bibitem[Lipovetsky \& Conklin(2001)Lipovetsky and
  Conklin]{RePEc:wly:apsmbi:v:17:y:2001:i:4:p:319-330}
Lipovetsky, S. and Conklin, M.
\newblock Analysis of regression in game theory approach.
\newblock \emph{Applied Stochastic Models in Business and Industry},
  17\penalty0 (4):\penalty0 319--330, October 2001.

\bibitem[Long(2022)]{Quan_Long_2022}
Long, Q.
\newblock Multimodal information gain in {B}ayesian design of experiments.
\newblock \emph{Computational Statistics}, 37:\penalty0 865--885, 2022.

\bibitem[Lundberg \& Lee(2017)Lundberg and Lee]{NIPS2017_8a20a862}
Lundberg, S.~M. and Lee, S.-I.
\newblock A unified approach to interpreting model predictions.
\newblock In Guyon, I., Luxburg, U.~V., Bengio, S., Wallach, H., Fergus, R.,
  Vishwanathan, S., and Garnett, R. (eds.), \emph{Advances in Neural
  Information Processing Systems}, volume~30. Curran Associates, Inc., 2017.

\bibitem[Moritz et~al.(2016)Moritz, Nishihara, and Jordan]{pmlr-v51-moritz16}
Moritz, P., Nishihara, R., and Jordan, M.
\newblock A linearly-convergent stochastic l-bfgs algorithm.
\newblock In Gretton, A. and Robert, C.~C. (eds.), \emph{Proceedings of the
  19th International Conference on Artificial Intelligence and Statistics},
  volume~51 of \emph{Proceedings of Machine Learning Research}, pp.\  249--258,
  Cadiz, Spain, 09--11 May 2016. PMLR.

\bibitem[Owen(2014)]{owen:SISV}
Owen, A.~B.
\newblock Sobol' indices and shapley value.
\newblock \emph{SIAM/ASA Journal on Uncertainty Quantification}, 2\penalty0
  (1):\penalty0 245--251, 2014.

\bibitem[Ribeiro et~al.(2016)Ribeiro, Singh, and Guestrin]{LIME}
Ribeiro, M.~T., Singh, S., and Guestrin, C.
\newblock why should i trust you explaining the predictions of any classifier.
\newblock In \emph{Proceedings of the 22nd ACM SIGKDD international conference
  on knowledge discovery and data mining}, pp.\  1135–1144, 2016.

\bibitem[Saltelli et~al.(2010)Saltelli, Annoni, Azzini, Campolongo, Ratto, and
  Tarantola]{SALTELLI2010259}
Saltelli, A., Annoni, P., Azzini, I., Campolongo, F., Ratto, M., and Tarantola,
  S.
\newblock Variance based sensitivity analysis of model output. design and
  estimator for the total sensitivity index.
\newblock \emph{Computer Physics Communications}, 181\penalty0 (2):\penalty0
  259--270, 2010.
\newblock ISSN 0010-4655.

\bibitem[Sobol(2001)]{SOBOL2001271}
Sobol, I.
\newblock Global sensitivity indices for nonlinear mathematical models and
  their monte carlo estimates.
\newblock \emph{Mathematics and Computers in Simulation}, 55\penalty0
  (1):\penalty0 271--280, 2001.
\newblock ISSN 0378-4754.
\newblock The Second IMACS Seminar on Monte Carlo Methods.

\bibitem[Tsai et~al.(2023)Tsai, Yeh, and Ravikumar]{JMLR:v24:22-0202}
Tsai, C.-P., Yeh, C.-K., and Ravikumar, P.
\newblock Faith-shap: The faithful shapley interaction index.
\newblock \emph{Journal of Machine Learning Research}, 24\penalty0
  (94):\penalty0 1--42, 2023.

\bibitem[Wu et~al.(2017)Wu, Poloczek, Wilson, and Frazier]{NIPS2017_64a08e5f}
Wu, J., Poloczek, M., Wilson, A.~G., and Frazier, P.
\newblock {B}ayesian optimization with gradients.
\newblock In Guyon, I., Luxburg, U.~V., Bengio, S., Wallach, H., Fergus, R.,
  Vishwanathan, S., and Garnett, R. (eds.), \emph{Advances in Neural
  Information Processing Systems}, volume~30. Curran Associates, Inc., 2017.

\bibitem[Yan et~al.(2017)Yan, Yin, and Sarkar]{NIPS2017_dfeb9598}
Yan, B., Yin, M., and Sarkar, P.
\newblock Convergence of gradient {EM} on multi-component mixture of gaussians.
\newblock In Guyon, I., Luxburg, U.~V., Bengio, S., Wallach, H., Fergus, R.,
  Vishwanathan, S., and Garnett, R. (eds.), \emph{Advances in Neural
  Information Processing Systems}, volume~30. Curran Associates, Inc., 2017.

\end{thebibliography}
\end{document}